\documentclass[runningheads]{llncs}
\usepackage{graphicx}
\usepackage{hyperref}
\usepackage{algorithm}
\usepackage{comment}
\usepackage{algpseudocode}
\usepackage{amsmath}
\usepackage{todonotes}
\usepackage{lastpage}  

\begin{document}
\title{IntelliMove: Enhancing Robotic Planning with Semantic Mapping}

\author{Fama NGOM\inst{1}\inst{,}\inst{3}, Huaxi (Yulin) ZHANG\inst{2}, Lei ZHANG\inst{1}, Karen GODARY-DEJEAN\inst{3}, Marianne HUCHARD\inst{3}}

\authorrunning{F. Author et al.}

\institute{S.A.S. euroDAO, 161 rue Ada, 34095 Montpellier cedex 5, France  \email \\ {\{fama.ngom,lei.zhang\}@eurodao.com} \and Université de Picardie Jules Verne, 
48 Rue d'Ostende, 02100 Saint Quentin, France  \email \\{yulin.zhang@u-picardie.fr} \and LIRMM, CNRS, Université de Montpellier, Montpellier, France 
\email \\ {\{karen.godary-dejean,marianne.huchard\}@lirmm.fr}
}

\maketitle              
\begin{abstract}

Semantic navigation enables robots to understand their environments beyond basic geometry, allowing them to reason about objects, their functions, and their interrelationships. In semantic robotic navigation, creating accurate and semantically enriched maps is fundamental. Planning based on semantic maps not only enhances the robot's planning efficiency and computational speed but also makes the planning more meaningful, supporting a broader range of semantic tasks.

In this paper, we introduce two core modules of IntelliMove: IntelliMap, a generic hierarchical semantic topometric map framework developed through an analysis of current technologies strengths and weaknesses, and Semantic Planning, which utilizes the semantic maps from IntelliMap. We showcase use cases that highlight IntelliMove's adaptability and effectiveness. Through experiments in simulated environments, we further demonstrate IntelliMove's capability in semantic navigation.

\end{abstract}

\section{Introduction}

Recent advancements in artificial intelligence have significantly shifted the focus towards augmenting the autonomy of robots by integrating high-level information into robotic systems. This shift has particularly influenced the domain of mobile robotics, where embedding semantic data into navigational functions has catalyzed the development of semantic navigation. Semantic navigation is defined as a system that incorporates semantic information into environmental representations, leveraging this data during robot localization and navigation processes to enrich understanding and functionality~\cite{Kostavelis2015,Wang2018Visual,Barber2018Mobile}.

\let\thefootnote\relax\footnotetext{This research was funded by the EUROCLUSTER DREAM Project under the sub-project \textit{IntelliMove} and by France BPI Innovation under Grant No. \textit{0208161/00}.}

Semantic mapping plays a pivotal role in robotics and autonomous systems, enhancing environmental models with high-level abstract elements that effectively bridge the gap between human understanding and robotic interpretation. The development of hierarchical semantic maps, integrating metric, topological, and semantic layers, is seen as a key advancement. The literature on hierarchical representations for semantic navigation, including hierarchical topometric models~\cite{Kuipers2000,Hiller2019,He2021} and 3D scene graphs~\cite{Armeni2019a,Kim2020,gu2023conceptgraphs}, reveals a trade-off between operational efficiency and semantic richness. Thus, a primary challenge remains: finding a balance between achieving computational efficiency and incorporating rich semantic details into map representations for effective semantic navigation.

The IntelliMove framework introduced in this paper comprises two main components: IntelliMap and Semantic Planning. IntelliMap is a hierarchical semantic topometric map framework that combines metric-semantic and topological-semantic data within a multi-layered structure to balance computational efficiency with environmental comprehension. Semantic Planning uses these enriched maps to enable robots to develop and execute contextually relevant navigation and task strategies, adapting to dynamic environmental changes. This planning process includes traditional graph-based pathfinding integrated with advanced semantic analysis, enhancing path generation from the start to designated goals, which could be specific objects or rooms, thus improving flexibility and efficiency.

The paper is organized as follows: Section 2 reviews related work, Section 3 describes the IntelliMap framework and its implementation, Section 4 explains the semantic planning algorithm, Section 5 details experiments and evaluations, and Section 6 concludes with future work directions.

\section{Related Work}
\subsubsection{Semantic Mapping.}
Within the present research context, two primary categories of semantic topometric maps have emerged as prominent. \\  \textit{1) Hierarchical Topometric Map.} Predominantly based on hierarchical topometric maps~\cite{Kuipers2000,Hiller2019,He2021}, this mapping approach utilizes inference methods on metric maps but lacks object-centric semantic details. It only recognizes basic categories such as rooms, corridors, and doors, showing limitations in early research works due to the lack of semantic scene understandings. \\ \textit{2) 3D Scene Graph.} To improve upon these maps, the 3D scene graph technique~\cite{Armeni2019a,Kim2020,hughes2022hydra,wu2021scenegraphfusion} provides an all-encompassing representation of the environment. It integrates a 3D semantic mesh layer, object layer, complex topological map, room layer, and building layer into the mapping process, requiring significant computational resources~\cite{Rosinol2020,Rosinol2021}.

Each of the mentioned map types has its own limitations, highlighting the need for a map type that balances efficiency, detail, and semantic understanding. In this paper, we proposes IntelliMap, a  hierarchical semantic topometric mapping framework that seamlessly integrates metric-semantic and topological-semantic representations in a balanced manner.

\subsubsection{Semantic Planning.}

Recent research has proposed methodologies for generating path planning that integrate semantic information of the environment. For instance, \cite{kremer2023s} introduces S-Nav, a semantic-geometric planner that enhances the performance of geometric planning through a hierarchical architecture. It exploits semantic information concerning room boundaries and doorways but does not assign semantic labels to rooms based on specific functions such as kitchens or offices. Similarly, \cite{serdel2023smana} presents the SMaNa navigation stack, which comprises an online 2.5D semantic navigation graph builder and a weighted A* pathfinder. While \cite{achat2022path} utilizes a semantic map consisting of three layers: a costmap, a conventional exploration grid, and a binary grid that records observations of distinct semantic categories. These works \cite{chaplot2020object,fukushima2022object,majumdar2022zson} generate paths to the designated semantic object goals based on object semantic map. However, they lack the ability to consider a higher-level semantic abstraction of the room environment, which would be beneficial for improved planning and decision-making. Finally,  \cite{sun2019semantic} conducts indoor path planning by utilizing a methodology that employs three layers for a 2D floor map, integrating predefined semantic encoding for objects, obstacles, dynamic entities, and room properties

As mentioned in the previous section, hierarchical semantic mapping integrated with metric maps is crucial not only for semantic planning but also for maintaining the precision of navigation. The works mentioned often focus on a single semantic layer of the environment, with or without a metric map. This limitation restricts their ability to achieve deeper semantic planning.

\section{IntelliMap: Semantic Mapping}
This section presents our framework, IntelliMap, a hierarchical semantic topometric mapping framework and one of its implementation.

\subsection{Framework}
\label{sec:overview}
\begin{figure*}[!thpb]
  \centering
  \includegraphics[width=0.98\textwidth]{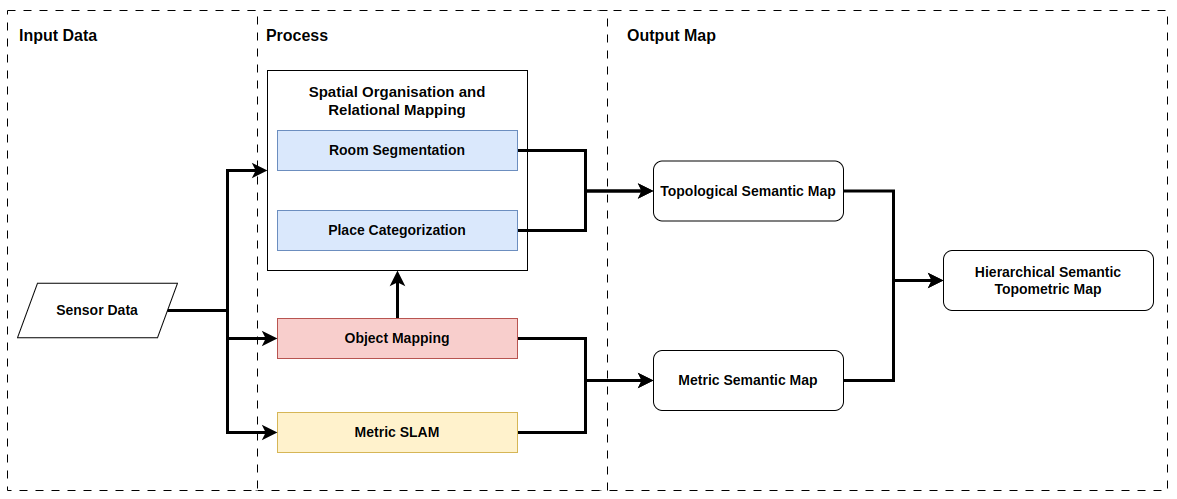}
  \caption{IntelliMap framework for semantic robot navigation. This figure illustrates the core components of the IntelliMap framework, designed to generate hierarchical semantic topometric maps for robot navigation in diverse environments. }
  \label{fig:general_IntelliMap_framework}
\end{figure*}

The IntelliMap framework in figure~\ref{fig:general_IntelliMap_framework} provides a robust approach to creating semantic maps for autonomous robotic navigation and decision-making across various settings. It starts with data collection via multiple sensors and utilizes a metric SLAM system to produce a geometrically accurate map. An Object Mapping component then integrates this data to identify and classify objects, organizing them spatially within the metric map to form a metric-semantic map or developing a separate topological semantic layer for specific needs. The process culminates in spatial organization and relational mapping, achieving a comprehensive topological semantic map.

\paragraph{Input Data: sensor data.}
\label{sec:inputdata}
The IntelliMap framework primarily utilizes camera inputs to capture essential visual information. 
Additional sensors like LiDAR and IMUs can be integrated for enhanced perception and accuracy. 

\paragraph{Map Construction Process.}
\label{sec:process}
\begin{itemize}
    \item \textbf{Metric SLAM.} In our IntelliMap framework, metric Simultaneous Localization and Mapping (SLAM)  module serves a dual purpose. First, it constructs a foundational metric map that captures the physical layout and geometry of the environment. Second, it continuously determines the robot's pose (position and orientation) within the constructed metric map. 
    \item \textbf{Object Mapping.} The Object Mapping module, integral to the SLAM system, captures and localizes objects using the robot's pose data. It transitions raw data into semantic insights, enabling interaction and navigation within a semantically rich environment. This module not only recognizes and locates distinct objects within visual data but also assigns precise positions relative to the metric map. It plays a crucial role in spatial organization by categorizing places based on the objects' presence and distribution. Object information can be integrated into the metric semantic map, where objects are directly incorporated into the metric map, or in a separate object map layer.
    \item \textbf{Room Segmentation.} The room segmentation algorithm can identify discernible environmental features that differentiate one room from another, such as physical boundaries (walls, floor-level variations), or object clusters. Approaches to this segmentation process are detailed on earlier studies.
    \item \textbf{Place Categorization.}  Without place categorization, the map might simply show rectangular rooms labeled as “Room 1”, “Room 2”, and so on. Thanks to place categorization, these rooms are identified as “Office”, “Living Room”, each carrying a specific meaning and context. Various approaches to achieve place categorization include ontology-based or LLM-based.
\end{itemize}

\paragraph{Output Maps.} 
The previous two steps generate two separate maps: a metric semantic map and a semantic topological map. While each map can be used for semantic robot navigation, our goal is to create a unified map that leverages both.
\begin{itemize}
    \item \textbf{Metric Semantic Map:} Combines metric SLAM data with semantic object mapping to detail both the environment's geometry and its object labels.
    \item \textbf{Topological Semantic Map:} Arises from Spatial Organisation and Relational Mapping, illustrating room connectivity and their functional relationships, useful for abstracted navigation and planning.
    \item \textbf{Hierarchical Semantic Topometric Map:} Integrates topological and metric information in a layered, hierarchical format, enhancing environmental representation. Its adaptable structure supports customization for specific applications, merging precise spatial details with a broader contextual understanding of space and object interactions.
\end{itemize}

\subsection{A Three-Layered Semantic Map Based on Costmap}

\begin{figure*}[!tpb]
  \centering
  \includegraphics[width=0.98\textwidth]{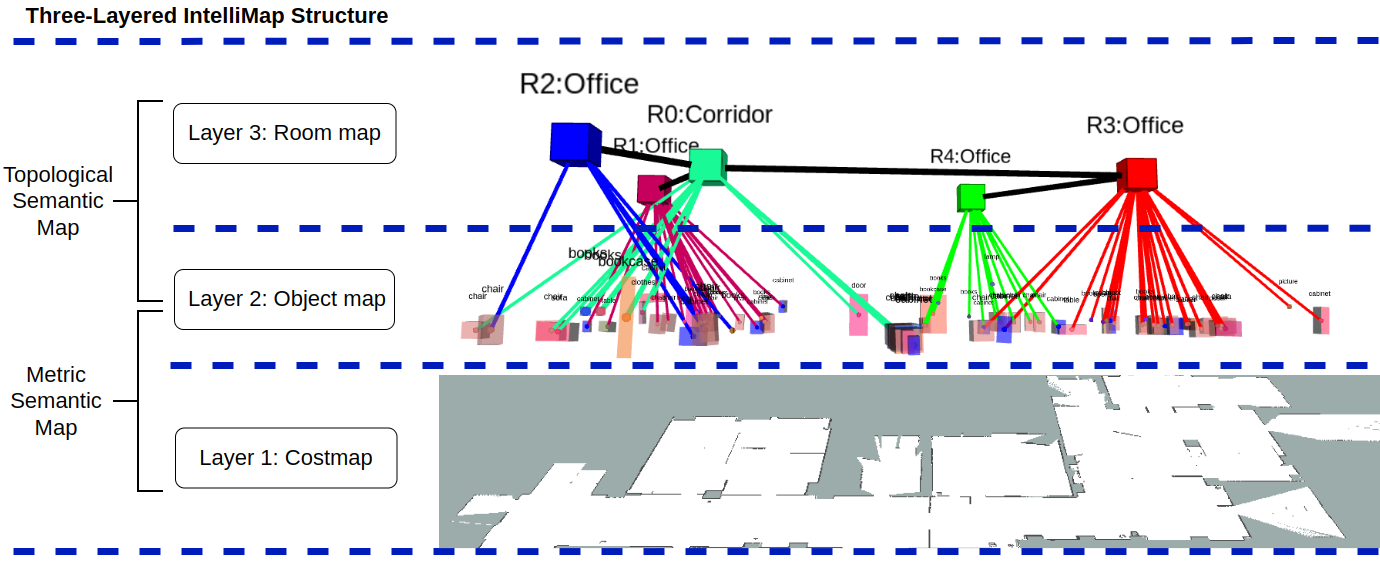}
  \caption{Three-Layered IntelliMap in a simulated office environment}
  \label{fig:3layer_IntelliMap}
\end{figure*}

This section introduces a three-layered semantic map integrating the IntelliMap framework to improve semantic robot navigation and spatial understanding in varied environments. The layers include Metric, Object, and Room categories. Figure~\ref{fig:3layer_IntelliMap} displays the IntelliMap output in a simulated office setting, utilizing the uHumans2 dataset~\cite{Rosinol2021}. This photo-realistic Unity simulation, enriched with diverse objects and spaces like desks, chairs, and corridors, serves as a preliminary evaluation of the framework's adaptability and effectiveness in complex settings. The three-layered structure significantly enhances navigation in indoor environments, with detailed results.

Our proposed IntelliMap framework, tailored for semantic robot navigation in diverse indoor environments utilizes data from various modules to construct a three-layered map (Figure~\ref{fig:3layer_IntelliMap}). 
\begin{itemize}
    \item \textbf{Metric Layer.} The Metric Map, generated through the SLAM process, encapsulates various spatial representations of the environment, such as costmaps, octomaps, or grid maps. 
    \item \textbf{Object Layer.} Derived from the Object Mapping module, this layer builds upon the Metric Map by identifying and incorporating objects within the environment. It features identifiable objects, each mapped with respect to its location and orientation, overlaying the Metric Map with detailed object information.
    \item \textbf{Room Layer.} The highest abstraction layer emerges from the place categorization and room segmentation modules. 
    By providing a topological overview of space categorization and room segmentation, this layer facilitates advanced decision-making and planning.
\end{itemize}
The metric layer and the object layer collectively form a metric semantic map, but are separated into distinct layers. The semantic topological map, serving as the third layer, merges with the previous layers to create the three-layered semantic map.

In the three-layered semantic map, semantic information is organized within a graph. The object layer represents objects as nodes, while the room layer assigns nodes to specific rooms. Graph edges denote physical connections between rooms, and connections between object and room layers indicate the presence of specific objects within their respective rooms.

\section{Semantic Planning Based on IntelliMap}

In this section, we present a semantic planning based on the IntelliMap introduced previously. By leveraging semantic mapping, robots achieve a deeper comprehension of their environments, enriched with semantic significance. This advanced paradigm in robotic navigation transcends traditional methods by incorporating extensive semantic information into navigation systems. Our approach to semantic planning not only utilizes fine-grained semantic details, such as objects, but also incorporates coarse-grained semantic information, such as rooms. This dual-level granularity enhances the robot's ability to navigate and interact within complex environments effectively.

\subsection{Overview}

\paragraph{Semantic planning definition.}

In this paper, we discuss semantic path planning, which combines graph-theoretical constructs with real-world navigation by leveraging IntelliMap's semantic graph. Nodes in this graph represent elements like rooms and objects, while edges, can be weighted by distance and energy, also encode semantic relationships between these elements. This integration not only enhances the robot's understanding of its environment but also optimizes navigation paths contextually and spatially. Consequently, the robot’s movements and tasks align more effectively with the environment's semantic structure, improving task performance efficiency.

\begin{figure}[tpb]
    \centering
    \includegraphics[width=0.98\textwidth]{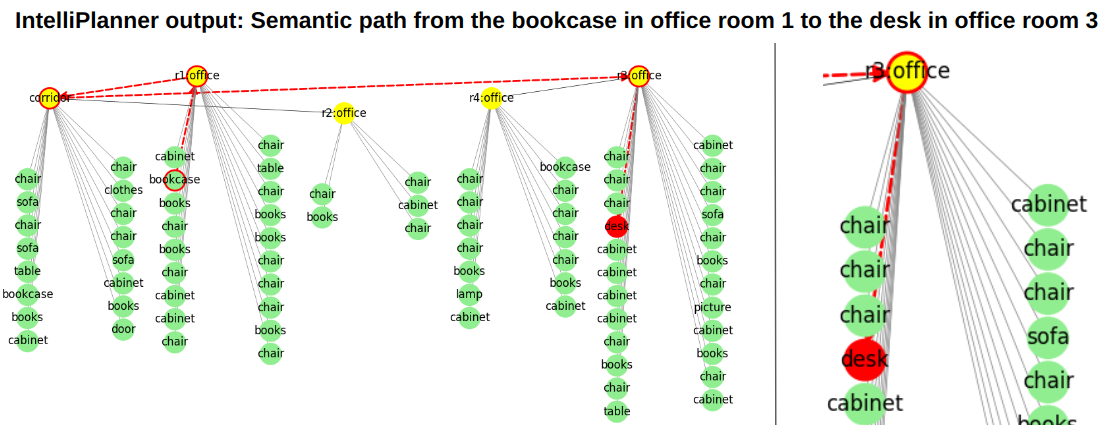}
    \caption{ Example of a red semantic path generated in an office environment, navigating from the bookcase in office room 1 to a desk in office room 3.}
    \label{fig:intelliplanner_path}
\end{figure}

\begin{algorithm}
\caption{Enhanced Semantic Planning with IntelliMove}
\label{algo:planning}
\begin{algorithmic}[1] 
\State \textbf{Inputs:} $semantic\_map$, $start$, $goal$
\State \textbf{Output:} $best\_path$
\Procedure{IntelliMoveSemanticPlanning}{$semantic\_map$, $start$, $goal$}
    \State $best\_path \gets \text{empty path}$
    \State $goal\_state \gets \Call{FindGoalState}{semantic\_map, goal}$
    \If{$goal\_state$ is empty}
        \State \textbf{Discovery Mode:}
        \State $goal\_LLM \gets \Call{GoalLLMResponse}{semantic\_map, goal}$
        \State $best\_path \gets \Call{Dijkstra}{semantic\_map, start, goal\_LLM}$
        \State \textbf{return} $best\_path$
    \ElsIf{length of $goal\_state = 1$}
        \State \textbf{Targeted Navigation Mode:}
        \State $best\_path \gets \Call{Dijkstra}{semantic\_map, start, goal}$
        \State \textbf{return} $best\_path$
    \Else
        \State \textbf{Multi-target Exploration Mode:}
        \State Initialize $min\_path\_length \gets \infty$
        \For{each $node$ in $goal\_state$}
            \State $path \gets \Call{Dijkstra}{semantic\_map, start, node}$
            \If{$path$ is not None and $path.length() < min\_path\_length$}
                \State $best\_path \gets path$
                \State $min\_path\_length \gets path.length()$
            \EndIf
        \EndFor
        \State \textbf{return} $best\_path$
    \EndIf
\EndProcedure
\end{algorithmic}
\end{algorithm}

\paragraph{Example.}
An example of semantic planning in an office environment is illustrated in Figure~\ref{fig:intelliplanner_path}, showing a path from a bookcase in office room 1 to a desk in office room 3. This example highlights our semantic planning capabilities for navigating complex semantic contexts. For example, when a robot is tasked with moving from an office to a conference room, our algorithm generates a path that includes all necessary intermediate rooms or nodes. This strategic navigation plan optimizes the route based on semantic relevance and travel costs, thus enhancing planning efficiency in large-scale environments compared to traditional metric maps.

\subsection{Algorithm of Planning}

The semantic planning process, detailed in Algorithm~\ref{algo:planning}, involves assessing the type of goal, integrating semantic data, and executing optimal path calculations to ensure that the robot's navigation is both effective and adaptively responsive to the environment's semantic attributes.

\paragraph{Path Input.}
Our semantic planning algorithm introduces a sophisticated approach that interprets and interacts with the environment, utilizing the following structured inputs to facilitate intelligent pathfinding:

\begin{itemize}
\item {Semantic\_map}: Represents the navigational space, organized into nodes that denote important locations such as rooms or objects. Edges connect rooms and delineate feasible paths, facilitating navigation within the environment. These edges also capture the semantic relationships between rooms and objects, with each edge can be weighted by factors such as distance, time, etc.
\item {Start Node}: Denotes the initial position of the robot within the semantic map, serving as the start point of the navigation task.
\item {Target Goal}: Specifies the target endpoint for the navigation, which could be either a specific object or a room.
\end{itemize}

\paragraph{Path Output.}
In the semantic planning, the output path plays an important role as the tangible result of the semantic planning process. This path represents the calculated route that the robot will follow to reach its designated goal. Below, the nature and significance of the output path are detailed, emphasizing its role in practical robot navigation.

The path is an ordered list of nodes starting from the initial position (start node) and ending at the target destination (goal node). Each node corresponds to a significant location within the environment, such as a specific room or object, defined in the semantic map.
The chosen path is the most efficient route calculated based on the weighted edges of the graph.
Each segment of the path is contextually relevant, reflecting the robot’s understanding of the environment’s semantic layout. This includes navigating through areas that are semantically tagged with specific functionalities or roles.

\paragraph{Process.}
The semantic planning process is meticulously outlined in the algorithm~\ref{algo:planning}, showing how it determines the most efficient route from the start node to the destination. This process involves assessing the type of goal, integrating semantic data, and executing optimal path calculations. By doing so, the algorithm ensures that the robot's navigation is both effective and adaptively responsive to the environment's semantic attributes.
The algorithm orchestrates sophisticated semantic planning by leveraging a structured approach based on the nature of the navigation goal.

Operational phases of the semantic planning algorithm are:
\begin{itemize}
    \item \textbf{Goal Evaluation and Path Planning}: The algorithm distinguishes between three types of navigation mode: Discovery Mode, Targeted Navigation Mode and Multi-target Exploration Mode.
        \begin{itemize}
            \item \textbf{Discovery Mode}: When the destination object is not predefined in the map, the algorithm extracts room attributes to understand the contextual setup of the environment. It then employs Large Language Model(LLM) to identify the most probable locations for the object, based on the semantic attributes of existing rooms within the semantic map. Once these probable locations are identified, Dijkstra’s algorithm is applied to calculate the most efficient routes to these rooms, ensuring that the navigation is directed towards the most likely locations as predicted by the Large Language Model (LLM). In our implementation, we employed the ChatGPT 3.5 large language model.
            
            \item \textbf{Targeted Navigation Mode}: The target point is singular. For navigation to a known destination, relevant nodes are identified. Subsequently, Dijkstra’s algorithm calculates the shortest paths from the start node to these identified targets, optimizing the route based on the weights assigned to the edges in the graph.
            
            \item \textbf{Multi-target Exploration Mode}: If there are multiple target points in the map, the robot will choose to explore the nearest target point.
        \end{itemize}
        
    \item \textbf{Path Execution}: the algorithm is designed to manage navigation efficiently by automatically selecting the optimal route.
 It evaluates and prioritizes paths based on their feasibility and proximity, thereby directing the robot to the most accessible location effectively. 
   
\end{itemize}

\section{Experimentation}
This section presents the findings of our comprehensive evaluation of our proposed IntelliMap framework and Semantic Planning in  office environments. 

\subsection{Evaluation}
\paragraph{Semantic Information Coverage.}
Table~\ref{tab:layer_configuration} provides a comparative analysis of the structure of IntelliMove's two components: IntelliMap and Semantic Planning, in comparison to other studies.
This analysis evaluates various features, including metric map representation, object and room representation, and object-based and room-based semantic planning. IntelliMove demonstrates the most extensive coverage across these features.

\begin{table*}[tpb]
\centering
\begin{tabular}{|p{3.0cm}|p{2.0cm}|p{1.4cm}|p{1.4cm}|p{1.5cm}|p{1.5cm}|}
\hline
\textbf{Method} & \textbf{Metric Map} & \multicolumn{2}{c|}{\textbf{Semantic Map}} & \multicolumn{2}{c|}{\textbf{Semantic Planning}} \\\cline{3-4} \cline{5-6}
 &  & \textbf{Object}  & \textbf{Room} & \textbf{Object Based} & \textbf{Room Based} \\ \hline
IntelliMove & Costmap & Yes & Yes & Yes & Yes \\ \hline
Zimmerman et al., 2023~\cite{zimmerman2023iros} & Costmap & Yes & No & No & No \\ \hline
Hughes et al., 2022~\cite{hughes2022hydra} & Mesh & Yes & Yes & No & No\\ \hline
He et al., 2021~\cite{He2021}  & Pointcloud & No & Yes & No & No \\ \hline
Hiller et al., 2019~\cite{Hiller2019}  & Costmap & No  & Yes & No & No \\ \hline
Gemignani et al.,2016~\cite{gemignani2016living} & Costmap & Yes  & Yes & No & No \\ \hline
Paul et al.,2023~\cite{kremer2023s}  & Room contours and SDF & Yes  & Yes & No & Yes \\ \hline
Devendra et al.,2020~\cite{chaplot2020object}  & Costmap & Yes  & No & Yes & No \\ \hline
\end{tabular}
\caption{Comparative analysis of layer configurations and semantic planning in IntelliMove and existing methods}
\label{tab:layer_configuration}
\end{table*}

\paragraph{Computational Efficiency.}
The experiments were conducted on a Dell Precision 7550 workstation equipped with 64.0 GiB of memory, an Intel® Core™ i9-10885H CPU @ 2.40GHz x 16 processor, and an NVIDIA GeForce RTX 3000 Mobile/Max-Q graphics card. 

We evaluated the semantic planning algorithm's performance focusing on its runtime and success rate. Runtime is defined as the duration from algorithm initiation to result generation, while the success rate measures the likelihood of accurately generating a path. In 50 tests within a simulated office environment, the average success time for existing targets was 7 milliseconds, with a maximum runtime of 10 milliseconds. The algorithm achieved a 99\% success rate for known targets and 55\% for goal discovery scenarios.

\subsection{Discussion}
Based on the experimental tests conducted, we can conclude that our algorithm provides a framework for semantic robot navigation, characterized by its efficiency, adaptability, and deep integration of semantic context:
\begin{itemize}
    \item \textbf{Efficiency and Precision}: It incorporates Dijkstra’s algorithm to ensure accurate and optimal pathfinding, continuously optimized through real-time data adjustments.
    \item \textbf{Adaptability}: Designed to handle both specific navigational targets and exploratory goals, the semantic planning seamlessly adapts to a wide range of operational scenarios, enhancing its functionality in complex environments.
    \item \textbf{Semantic Awareness}: The system’s use of AI for semantic data analysis guarantees navigation strategies that are deeply aware of the context, significantly improving robot interactions within their operational environment.
\end{itemize}
   
Our semantic planning approach to semantic robot navigation merges traditional graph-based pathfinding techniques with advanced semantic analysis, enabling efficient and context-aware navigation. This dual capability ensures that robots equipped with this algorithm are not only capable of navigating through intricate settings but can also adjust their paths in response to changing  conditions and sophisticated mission requirements. 

\section{Conclusion}
In this paper, we present the IntelliMove framework, which comprises two fundamental components: IntelliMap, a hierarchical semantic topometric mapping framework, and Semantic Planning. IntelliMap is engineered to generate nuanced and adaptable maps tailored for semantic robot navigation. By harnessing existing methodologies like hierarchical topometric maps and 3D scene graph while mitigating their drawbacks (e.g., high computational overhead, limited scalability), IntelliMap offers a modular, flexible, and resilient solution.

Building upon the IntelliMap framework, semantic planning merges traditional pathfinding techniques with advanced semantic analysis to provide not just efficient but also context-aware navigation. This dual functionality ensures navigation strategies that are highly adaptable and precise, enabling robots to traverse intricate environments with a level of understanding akin to human perception. The successful deployment of IntelliMap and our semantic planning algorithm within the IntelliMove project underscores their efficacy. This paper illustrates how these frameworks facilitate swifter and more efficient navigation by integrating semantic insights at various stages of the process.

\bibliographystyle{IEEEtran}
\bibliography{IEEEabrv,semantic}

\begin{thebibliography}{10}
\providecommand{\url}[1]{#1}
\csname url@samestyle\endcsname
\providecommand{\newblock}{\relax}
\providecommand{\bibinfo}[2]{#2}
\providecommand{\BIBentrySTDinterwordspacing}{\spaceskip=0pt\relax}
\providecommand{\BIBentryALTinterwordstretchfactor}{4}
\providecommand{\BIBentryALTinterwordspacing}{\spaceskip=\fontdimen2\font plus
\BIBentryALTinterwordstretchfactor\fontdimen3\font minus \fontdimen4\font\relax}
\providecommand{\BIBforeignlanguage}[2]{{%
\expandafter\ifx\csname l@#1\endcsname\relax
\typeout{** WARNING: IEEEtran.bst: No hyphenation pattern has been}%
\typeout{** loaded for the language `#1'. Using the pattern for}%
\typeout{** the default language instead.}%
\else
\language=\csname l@#1\endcsname
\fi
#2}}
\providecommand{\BIBdecl}{\relax}
\BIBdecl

\bibitem{Kostavelis2015}
I.~Kostavelis and A.~Gasteratos, ``Semantic mapping for mobile robotics tasks: {A} survey,'' \emph{Robotics Auton. Syst.}, vol.~66, pp. 86--103, 2015.

\bibitem{Wang2018Visual}
L.~Wang, L.~Zhao, G.~Huo, R.~Li, Z.~Hou, P.~Luo, Z.~Sun, K.~Wang, and C.~Yang, ``Visual semantic navigation based on deep learning for indoor mobile robots,'' \emph{Complexity}, vol. 2018, 2018.

\bibitem{Barber2018Mobile}
R.~Barber, J.~Crespo, C.~G{\'o}mez, A.~C. Hern{\'a}mdez, and M.~Galli, ``Mobile robot navigation in indoor environments: Geometric, topological, and semantic navigation,'' in \emph{Applications of Mobile Robots}.\hskip 1em plus 0.5em minus 0.4em\relax IntechOpen, 2018.

\bibitem{Kuipers2000}
B.~Kuipers, ``The spatial semantic hierarchy,'' \emph{Artificial Intelligence}, vol. 119, no.~1, pp. 191--233, 2000.

\bibitem{Hiller2019}
M.~Hiller, C.~Qiu, F.~Particke, C.~Hofmann, and J.~Thielecke, ``Learning topometric semantic maps from occupancy grids,'' in \emph{2019 IEEE/RSJ International Conference on Intelligent Robots and Systems (IROS)}, 2019, pp. 4190--4197.

\bibitem{He2021}
Z.~He, H.~Sun, J.~Hou, Y.~Ha, and S.~Schwertfeger, ``Hierarchical topometric representation of 3d robotic maps,'' \emph{Auton. Robots}, vol.~45, no.~5, pp. 755--771, 2021.

\bibitem{Armeni2019a}
I.~Armeni, Z.-Y. He, A.~Zamir, J.~Gwak, J.~Malik, M.~Fischer, and S.~Savarese, ``3d scene graph: A structure for unified semantics, 3d space, and camera,'' in \emph{Proceedings of the IEEE/CVF international conference on computer vision}.\hskip 1em plus 0.5em minus 0.4em\relax Seoul, Korea (South): IEEE, 2019, pp. 5663--5672.

\bibitem{Kim2020}
U.-H. Kim, J.-M. Park, T.~jin Song, and J.-H. Kim, ``{3-D Scene Graph: A Sparse and Semantic Representation of Physical Environments for Intelligent Agents},'' \emph{IEEE Transactions on Cybernetics}, vol.~50, pp. 4921--4933, 2020.

\bibitem{gu2023conceptgraphs}
Q.~Gu, A.~Kuwajerwala, S.~Morin, K.~M. Jatavallabhula, B.~Sen, A.~Agarwal, C.~Rivera, W.~Paul, K.~Ellis, R.~Chellappa \emph{et~al.}, ``Conceptgraphs: Open-vocabulary 3d scene graphs for perception and planning,'' \emph{arXiv preprint arXiv:2309.16650}, 2023.

\bibitem{hughes2022hydra}
N.~Hughes, Y.~Chang, and L.~Carlone, ``Hydra: A real-time spatial perception system for 3d scene graph construction and optimization,'' \emph{Robotics: Science and Systems XVIII}, 2022.

\bibitem{wu2021scenegraphfusion}
S.-C. Wu, J.~Wald, K.~Tateno, N.~Navab, and F.~Tombari, ``Scenegraphfusion: Incremental 3d scene graph prediction from rgb-d sequences,'' in \emph{Proceedings of the IEEE/CVF Conference on Computer Vision and Pattern Recognition}, 2021, pp. 7515--7525.

\bibitem{Rosinol2020}
A.~Rosinol, A.~Gupta, M.~Abate, J.~Shi, and L.~Carlone, ``{3D Dynamic Scene Graphs: Actionable Spatial Perception with Places, Objects, and Humans},'' in \emph{Robotics: Science and Systems XVI}, M.~Toussaint, A.~Bicchi, and T.~Hermans, Eds., Oregon, USA, 2020.

\bibitem{Rosinol2021}
A.~Rosinol, A.~Violette, M.~Abate, N.~Hughes, Y.~Chang, J.~Shi, A.~Gupta, and L.~Carlone, ``{Kimera: From {SLAM} to spatial perception with 3D dynamic scene graphs},'' \emph{Int. J. Robotics Res.}, vol.~40, no. 12-14, pp. 1510--1546, 2021.

\bibitem{kremer2023s}
P.~Kremer, H.~Bavle, J.~L. Sanchez-Lopez, and H.~Voos, ``S-nav: Semantic-geometric planning for mobile robots,'' \emph{arXiv preprint arXiv:2307.01613}, 2023.

\bibitem{serdel2023smana}
Q.~Serdel, J.~Marzat, and J.~Moras, ``Smana: Semantic mapping and navigation architecture for autonomous robots,'' in \emph{Proceedings of the 20th International Conference on Informatics in Control, Automation and Robotics, {ICINCO} 2023, Rome, Italy, November 13-15, 2023, Volume 1}, G.~Gini, H.~Nijmeijer, and D.~P. Filev, Eds.\hskip 1em plus 0.5em minus 0.4em\relax {SCITEPRESS}, 2023, pp. 453--464.

\bibitem{achat2022path}
S.~Achat, J.~Marzat, and J.~Moras, ``Path planning incorporating semantic information for autonomous robot navigation,'' in \emph{19th International Conference on Informatics in Control, Automation and Robotics (ICINCO)}, 2022.

\bibitem{chaplot2020object}
D.~S. Chaplot, D.~P. Gandhi, A.~Gupta, and R.~R. Salakhutdinov, ``Object goal navigation using goal-oriented semantic exploration,'' \emph{Advances in Neural Information Processing Systems}, vol.~33, pp. 4247--4258, 2020.

\bibitem{fukushima2022object}
R.~Fukushima, K.~Ota, A.~Kanezaki, Y.~Sasaki, and Y.~Yoshiyasu, ``Object memory transformer for object goal navigation,'' in \emph{2022 International Conference on Robotics and Automation (ICRA)}.\hskip 1em plus 0.5em minus 0.4em\relax IEEE, 2022, pp. 11\,288--11\,294.

\bibitem{majumdar2022zson}
A.~Majumdar, G.~Aggarwal, B.~Devnani, J.~Hoffman, and D.~Batra, ``Zson: Zero-shot object-goal navigation using multimodal goal embeddings,'' \emph{Advances in Neural Information Processing Systems}, vol.~35, pp. 32\,340--32\,352, 2022.

\bibitem{sun2019semantic}
N.~Sun, E.~Yang, J.~Corney, and Y.~Chen, ``Semantic path planning for indoor navigation and household tasks,'' in \emph{Towards Autonomous Robotic Systems: 20th Annual Conference, TAROS 2019, London, UK, July 3--5, 2019, Proceedings, Part II 20}.\hskip 1em plus 0.5em minus 0.4em\relax Springer, 2019, pp. 191--201.

\bibitem{zimmerman2023iros}
N.~Zimmerman, M.~Sodano, E.~Marks, J.~Behley, and C.~Stachniss, ``{Constructing Metric-Semantic Maps Using Floor Plan Priors for Long-Term Indoor Localization},'' in \emph{IEEE/RSJ Intl.~Conf.~on Intelligent Robots and Systems (IROS)}, 2023.

\bibitem{gemignani2016living}
G.~Gemignani, R.~Capobianco, E.~Bastianelli, D.~D. Bloisi, L.~Iocchi, and D.~Nardi, ``Living with robots: Interactive environmental knowledge acquisition,'' \emph{Robotics and Autonomous Systems}, vol.~78, pp. 1--16, 2016.

\end{thebibliography}
\end{document}